\documentclass{article}

\usepackage{iclr2025_conference,times}

\iclrfinalcopy

\usepackage[table,xcdraw,dvipsnames]{xcolor}
\usepackage{subcaption}
\usepackage{amsmath,amssymb,amsfonts}
\usepackage{graphicx}
\usepackage{textcomp}
\usepackage{arydshln}
\usepackage{algorithm}
\usepackage{algorithmic}
\usepackage{booktabs}
\usepackage{multirow}
\usepackage{hyperref}

\title{Reducing Object Hallucination in LVLMs \\ via Emphasizing Image-negative Tokens}

\author{Meng Shen \\ Nanyang Technological University \\ \texttt{meng005@e.ntu.edu.sg}
\And
Minghao Wu \\ Monash University \\ \texttt{minghao.wu@monash.edu}
\And
Deepu Rajan \\ Nanyang Technological University \\ \texttt{ASDRajan@ntu.edu.sg}
}

\begin{document}

\maketitle
% Clear conference header for clean arXiv preprint
\lhead{}

\begin{abstract}
Object hallucination is a significant challenge that hinders the application of large vision-language models (LVLMs) in practice. We hypothesize that one possible origin of hallucination is the model's tendency to prioritize text generation over meaningful interaction with images. To explore this, we examine the generation process and categorize text tokens into three groups: image-positive, invariant, and negative, based on their visual dependence on input image tokens. Our analysis reveals that most generated tokens are minimally influenced by the image information. This suggests that during the model's training stage, more emphasis is placed on learning how to follow textual instructions, rather than extracting information from images. Based on this finding, we propose adjusting the training weights of different tokens depending on their visual dependence to control hallucination.  Additionally, we remove a portion of the training data that potentially contains more hallucinations as a data filtering strategy. Both methods achieve a reduction in hallucination without compromising response length or introducing additional computational costs during inference. We validate our methods across three LVLM variants, demonstrating the effectiveness and general applicability.
\end{abstract}

%-------------------------------------------------------------------------
\section{Introduction}

In recent years, large vision-language models (LVLMs) have integrated vision capabilities into language models, opening new avenues for AI applications \citep{llava, gpt4, beyer2024paligemma}. However, one of the most challenging issues in deploying LVLMs in real-world scenarios is the problem of hallucination. Hallucination occurs when the models generate plausible but inaccurate content that is not grounded in the image or not related to the provided text prompt. This issue can be particularly harmful in downstream tasks such as medical diagnostics \citep{DBLP:journals/corr/abs-2406-10185} and robotics \citep{DBLP:conf/corl/HuangWZL0023}, where the accurate extraction of image details is required. Therefore, addressing the hallucination problem is crucial for applying LVLMs in various practical applications.

To mitigate hallucination in LVLMs, several approaches have been proposed~\citep{opera, vcd, eos, liu2023mitigating}. However, these approaches often require additional instructional data, which are expensive to collect, or incur extra computational costs during inference. In this work, we aim to reduce hallucination during the training phase. While EOS \citep{eos} mitigates hallucination by training the model to stop generating redundant responses, which are often hallucinatory, it produces overly short answers that may not be desirable. We propose a method to reduce hallucination that does not require additional instructional data or increase computational costs during inference, while maintaining appropriate response lengths.

\begin{figure*}
    \centering
    \begin{subfigure}[t]{0.13\linewidth}
        \centering
        \raisebox{0.6cm}{%
            \includegraphics[width=\linewidth]{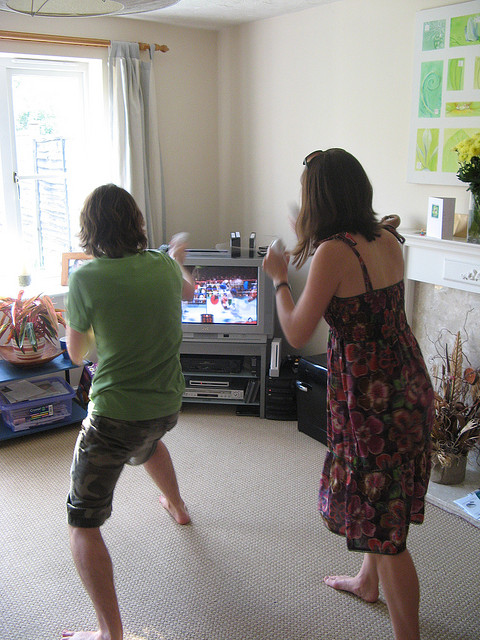}%
        }
    \end{subfigure}
    \hfill
    \begin{subfigure}[t]{0.13\linewidth}
        \centering
        \raisebox{0.6cm}{%
            \includegraphics[width=\linewidth]{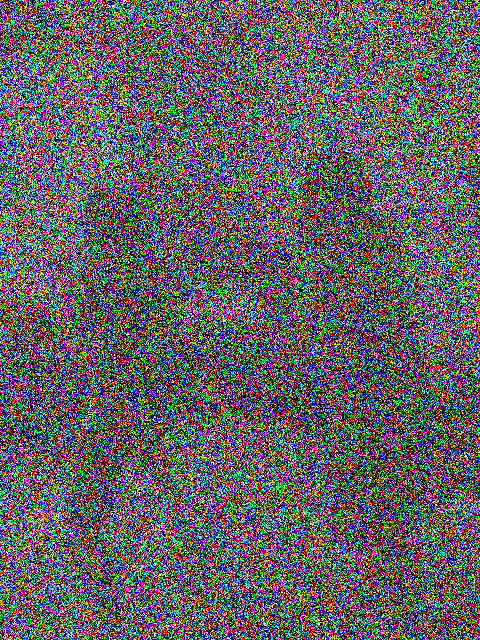}%
        }
    \end{subfigure}
    \hfill
    \begin{subfigure}[t]{0.72\linewidth}
        \centering
        \includegraphics[width=\linewidth]{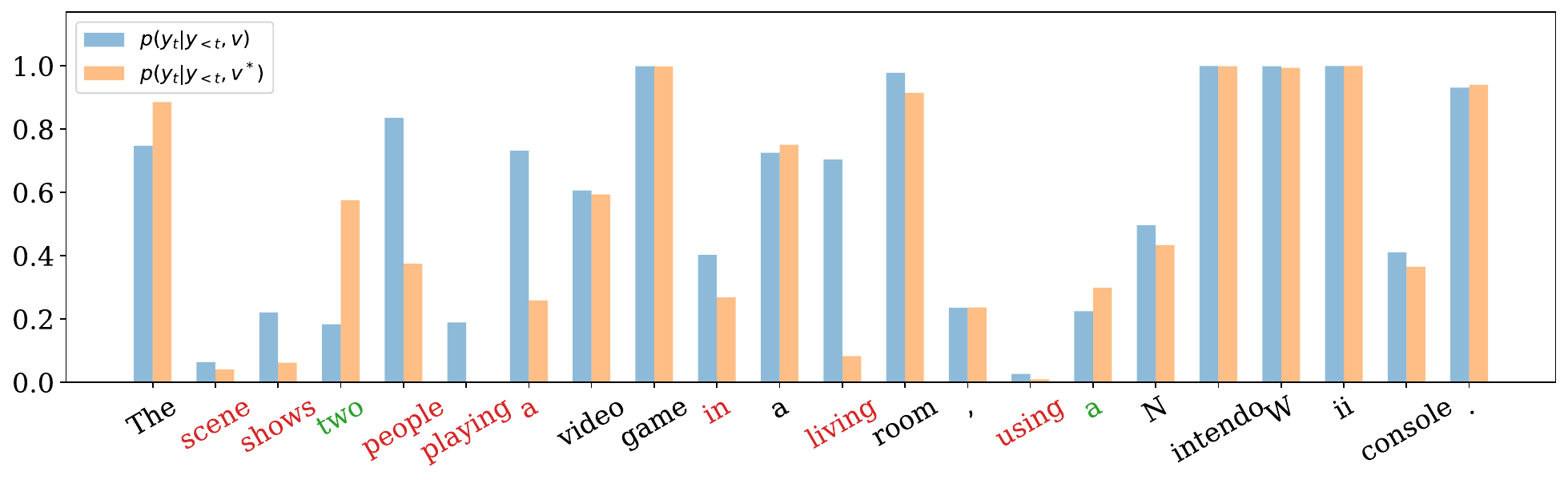}
    \end{subfigure}
    \caption{The left figure displays the input image and the middle figure is the noisy image with diffusion noise at time step 900. The model (fine-tuned LLaVA-v1.5-7b) is asked to generate a descriptive response given the image. The right figure shows the probability of each token $y_t$ when provided with a clean image  $v$  and a noisy image  $v^*$. The \textbf{image-positive}, \textbf{image-invariant}  and \textbf{image-negative} tokens are marked as {\color{BrickRed}red}, {\color{Black}black} and {\color{Green}green}.}
    \label{fig:prob_distribution_without_with_noise}
\end{figure*}

We begin by investigating the origins of hallucinations in LVLMs, focusing on the influence of images on the probability distribution of tokens in generated responses.  We propose a visual dependence metric by feeding LVLMs a clean image and a noisy image, and group the response tokens into three distinct categories based on this metric. \textbf{Figure \ref{fig:prob_distribution_without_with_noise}} illustrates the changes in token probabilities when a clean input image is replaced by a noisy one. In this paper, we demonstrate that the origins of hallucinations in generated text tokens vary based on their degree of correlation between image and text tokens.

To explore how different token types affect hallucination during the training stage, we propose scaling the loss of tokens according to their dependence on the image. Unlike traditional supervised fine-tuning of LVLMs, which treats all tokens equally, our approach adjusts the loss by suppressing or amplifying certain tokens throughout training. We find that emphasizing image-positive tokens during training results in both increased hallucinations and richer responses with more visual details.  In contrast, emphasizing image-negative tokens while suppressing image-positive and image-invariant tokens helps reduce hallucinations. Additionally, we implement a simple data filtering strategy to remove instructional samples based on their visual dependence. We find that removing a certain portion of instructions with high visual dependence effectively mitigates hallucinations.

\section{Related Works}

\paragraph{Causes of Hallucination} In the context of LVLMs, we focus on the image hallucination. The causes can be understood from several perspectives. \textbf{Training Data}: Misaligned visual instructional training data \citep{sun2023aligning} and data containing hallucinations \citep{liu2023mitigating} can introduce hallucination into the model during training. \textbf{Decreased Visual Information}: As the length of the generated response increases, the model is more likely to produce hallucinated content due to the gradual loss of visual information \citep{favero2024multi, eos, zhou2023analyzing}. \textbf{Language Prior}: Hallucination can occur when the model's language processing ability exceeds its vision processing ability, leading the model to overlook the image content \citep{zhou2023analyzing, vcd, han2022visual}. \textbf{Visual Encoder}: The current design of LVLMs often involves extracting hidden states from a pretrained vision model and feeding them into a language model. Therefore, when the visual representations are not good enough, the model is not able to generate consistent responses \citep{tong2024eyes, jiang2023clip}.

\paragraph{Mitigate Hallucination} The hallucination reduction methods can be categorized into three classes based on the stage they target: data preparation, training and inference. \textbf{Data curation} methods \citep{liu2023mitigating} focus on collecting high-quality data with less hallucination, ensuring that the trained model produces robust responses. \textbf{Training-based} methods \citep{eos, zhai2023halle} concentrate on manipulating the training process by either adjusting the generation length or training a plug-in module to control the level of hallucination. \textbf{Training-free/Inference} methods \citep{opera, vcd} aim to prevent the generation of hallucinated content through techniques such as contrastive decoding or rolling back when summary tokens are detected.

\section{Analysis of Hallucination in LVLMs}

Given an input image $v$, the generative LVLM $\theta$ produces a response in an auto-regressive manner. The probability of generating each token $y_t$ at time step $t$ is denoted as $p(y_t|y_{<t}, v)$, where $y_{<t}$ represents the preceding tokens, and $p$ is the probability of the current token $y_{t}$ conditioned on previous tokens $y_{<t}$ and visual input $v$. The images, as the conditional variants, are significant to hallucinations in LVLMs. In this section, we investigate how the input image affects the probability distribution and its relationship to hallucination.

\subsection{Visual Dependence}

To quantitatively assess the relationship between text tokens and input image, we propose \emph{visual dependence}, which reflects how generated text tokens are related with the input image by providing the model with an additional noisy image $v^*$.  The noisy image is created by applying Gaussian noise to the original image, similar to the forward diffusion process used in VCD \citep{vcd}. The strength of the applied Gaussian noise is controlled by a time step $T$, which ranges from 0 to 1000, where $T=0$ indicates clean image without any noise, and $T=1000$ indicates image is completely replaced by Gaussian noise. We set the time step to 900, which retains minimal visual information while still allowing the model to perceive some aspects of the image without being entirely blinded.
We represent the visual dependence of each token using the relative difference in token probability:

\begin{equation}
    d(y_t) = \frac{p(y_t|y_{<t}, v) - p(y_t|y_{<t}, v^*)}{\max[p(y_t|y_{<t}, v), p(y_t|y_{<t}, v^*)]}.
\end{equation}

The visual dependence value $d(y_t)$ for a token $y_t$ ranges between $-1$ and $1$. When $d(y_t)$ is significantly greater than $0$ (within the range \textbf{[0.25, 1]}), it indicates that the probability of generating this token decreases with the noisy image, suggesting that the token relies on the image content. Conversely, when $d(y_t)$ is significantly less than $0$ (within the range \textbf{[-1.0, -0.25)}), it means that this token is more likely to be generated even without image information, indicating a negative relationship with the visual content. As stated earlier, we refer to these two types of tokens as \textbf{image-positive} and \textbf{image-negative} tokens, respectively. Additionally, if a token has a $d(y_t)$ value that is close to 0 (within the range \textbf{[-0.25, 0.25)}), we classify it as an \textbf{image-invariant} token, as it is not significantly affected by the distorted image. For illustration, we provide an example in \textbf{Figure \ref{fig:prob_distribution_without_with_noise}} with noise step set at 900.

\subsection{Hallucinations in Generated Text Tokens}

\begin{figure}[t]
    \centering
    \begin{subfigure}[t]{0.45\textwidth}
        \centering
        \includegraphics[width=\linewidth]{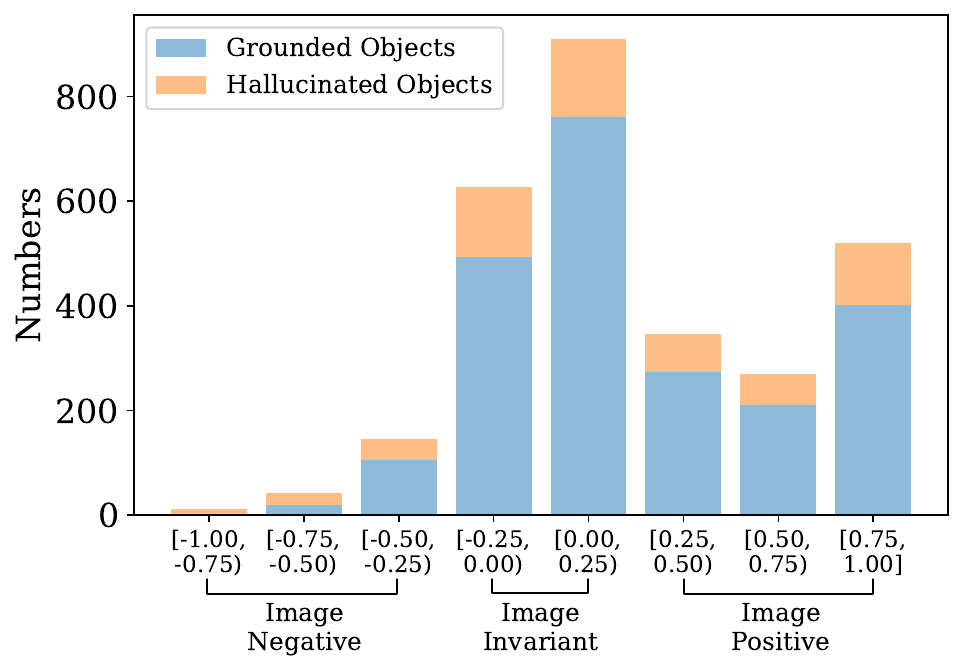}
        \caption{The number of grounded and hallucinated objects in different groups of tokens.}
        \label{fig:inspect_chair}
    \end{subfigure}%
    \hfill
    \begin{subfigure}[t]{0.5\textwidth}
        \centering
        \includegraphics[width=\linewidth]{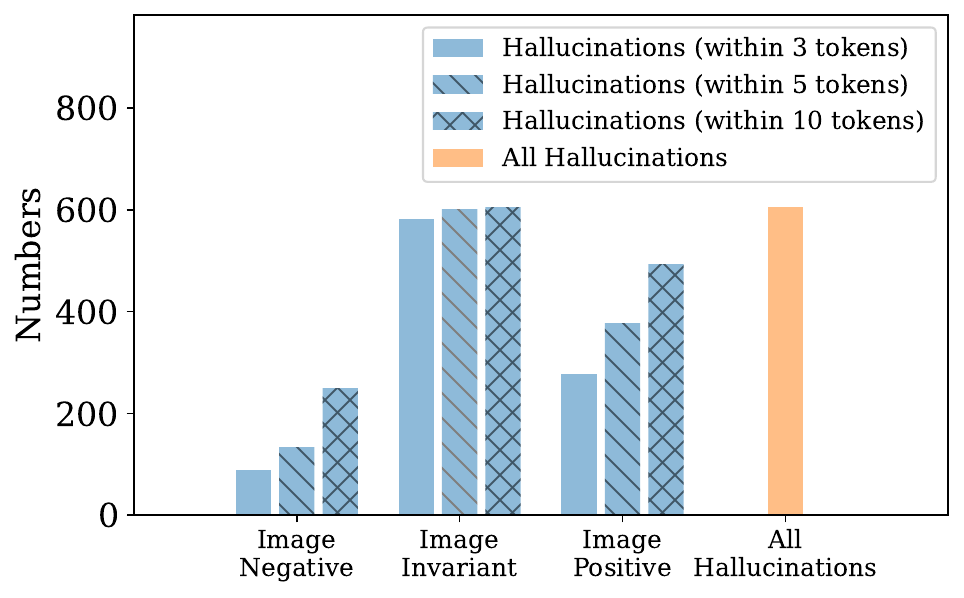}
        \caption{The numbers of hallucinations appear at different ranges of different tokens.}
        \label{fig:inspect_chair_co_occur}
    \end{subfigure}
\end{figure}

After quantifying the visual dependence of each token and classifying them as image-invariant, image-positive, or image-negative, we analyze the prevalence of hallucination within these different types of tokens. To achieve this, we adopt fine-tuned LLaVA-v1.5-7b \footnote{https://huggingface.co/liuhaotian/llava-v1.5-7b} to generate detailed descriptions for 500 randomly selected images from the COCO dataset \citep{cocodataset}. The prompts used is: "Please describe this image in detail.". We then extract both the grounded and hallucinated objects from the generated responses and count the occurrences within the different groups of tokens based on their visual dependence. The results are presented at \textbf{Figure \ref{fig:inspect_chair}}. We then analyze the co-occurrence of hallucinations of different types of text tokens using the same 500 randomly selected images from the COCO dataset. The results are shown in the \textbf{Figure \ref{fig:inspect_chair_co_occur}}.

\label{sec:hallucination_in_text_tokens}

\textbf{Findings} As shown in \textbf{Figure \ref{fig:inspect_chair}}, image-invariant tokens contain the majority of both grounded and hallucinated objects. In the example from \textbf{Figure \ref{fig:prob_distribution_without_with_noise}}, the brand of console 'Nintendo Wii' is difficult or ambiguous to detect from the image. The first token, "N", and the last token, "console", are image-related, while the other tokens are totally image-invariant, as they often appear together as a combination. Moreover, \textbf{Figure \ref{fig:inspect_chair_co_occur}} shows that around 95\% of hallucinations occur within 3 tokens of an image-invariant token, and only approximately half of hallucinations appear within 3 tokens of an image-positive token. This suggests that hallucinations in image-invariant tokens are primarily caused by the language generation process rather than poor visual information extraction capabilities. Additionally, \textbf{Figure \ref{fig:inspect_chair}} demonstrates that there are more objects in image-positive tokens than in image-negative tokens, which aligns with our expectation that image information should be more positively correlated with image inputs. Based on these observations, we conclude the following: \textbf{(1)} Hallucinations are more frequent in and around image-invariant tokens, and they often appear as commonly paired object combinations. \textbf{(2)} Image-positive tokens are more likely to be associated with extracting useful information from the image content, whereas image-negative tokens are more often non-object terms.

The first finding reveals that the model generates hallucinated objects associated with commonly co-occurring objects, even without directly seeing the image. These tokens are mostly image-invariant as they are less affected by the image content. It suggests that we can suppress the image-invariant tokens during training to prevent this type of hallucination. This co-occurrence phenomenon has been discussed in \citep{zhou2023analyzing}, however, they do not connect it to visual dependence.  The second finding suggests that by emphasizing image-positive tokens during training, the model might be more inclined to focus on the actual image content. However, our experimental results in \textbf{Section \ref{sec:exp}} indicate that while emphasizing image-positive tokens does improve object recognition rates, it leads to an increase in hallucination rates.

\section{Adjust Loss Weights During Training}

Our analysis suggests that different caption tokens have varying impacts on hallucinations in LVLMs, where image-invariant tokens cause a significant influence on hallucinations which need to be suppressed during the training. Therefore, we propose adjusting the training loss weights for these tokens based on their visual dependence. We design two re-weighting strategies: one that up-scales the loss for image-negative tokens, and another that up-scales the loss for image-positive tokens. Each strategy targets different token types, allowing us to study their effects on hallucination. The re-weighting functions are defined as follows:

\begin{equation}
w_{\text{neg}}(y_t) = \begin{cases}
-d(y_t) & \text{if } d(y_t) <= 0, \\
0 & \text{if } d(y_t) > 0, \\
\end{cases}
\end{equation}

\begin{equation}
w_{\text{pos}}(y_t) = \begin{cases}
0 & \text{if } d(y_t) <= 0, \\
d(y_t) & \text{if } d(y_t) > 0, \\
\end{cases}
\end{equation}

where $w_{\text{neg}}$ is applied when emphasizing image-negative tokens,  while $w_{\text{pos}}$ is used for emphasizing image-positive tokens. It is worth noting that, in both strategies, the weights of image-invariant tokens are down-scaled. To redistribute the loss weights without altering the overall magnitude of the original loss, we re-normalize the weights using softmax with a temperature parameter $\tau$, controlling the sharpness of weight distributions, and derive the re-scaled loss objective:

\begin{equation}
    \mathcal{L}(y) = -\frac{1}{T}\sum_{t=1}^{T} T \frac{e^{\tau w(y_t)}}{\sum_{1}^{T} e^{\tau w(y_t)}}\log p(y_t|y_{<t}, v).
\end{equation}

With this re-scaling loss, for example, when using $w_{neg}$ as $w(y_t)$, the loss weights become greater than $1$ for image-negative tokens and less than $1$ for image-positive and image-invariant tokens. In our experiments, we find that excessively down-scaling the loss weight of the EOS token could affect generation lengths. Therefore, we ensure that the loss weight of the EOS token is never reduced below $1$ to avoid this issue. The computation of $d(y_t)$ is performed on-the-fly right before loss computation for every batch of data, and it is done in evaluation mode to save GPU memory. The computational cost can be found in Appendix~\ref{sec:exp_settings}. To allow the model to accurately assess visual dependence, we begin loss re-weighting after half of training by default. In the appendix, we study the effects of different temperatures, starting times for adjustment, and various noise time steps used in calculation of visual dependence.

\section{Experiments}
\label{sec:exp}

\subsection{Experiment Setup}

\paragraph{LVLMs and Datasets} We select three LVLMs variants with different parameter sizes: PaliGemma-3b-224 \citep{beyer2024paligemma}, LLaVA-v1.5-7b and 13b \citep{llava} and Bunny-v1.1-8B \citep{bunny}, to evaluate our methods. We use LLaVA-Detail 23k and LLaVA-Instruct 150k \citep{llava} for fine-tuning LLaVA models. Since PaliGemma-3b is a single-turn model, we use LLaVA-Detail 23k, which is also a single-turn dataset, to fine tune it. For Bunny-v1.1-8b, we use the subset from its instruction fine-tune dataset SVIT-Detail 71k \citep{zhao2023svit} to conduct training.

\paragraph{Benchmarks} We evaluate models using CHAIR \citep{chair} and FaithScore \citep{faithscore}, and two yes/no VQA benchmarks: POPE \citep{pope2023} and MME \citep{mme2023}. CHAIR identifies hallucinations by detecting objects not included in the ground-truth label set from COCO annotations. We report the sentence-level score $\text{CHAIR}_{S}$ and the instance-level score $\text{CHAIR}_{I}$. Additionally, we report the Recall score to evaluate object recognition rates. FaithScore evaluates the consistency of atomic facts extracted from the generated captions with the image content. We report both the instance-level and sentence-level FaithScore. In MME, we report the perception score $\text{MME}_{\text{P}}$ and cognition score $\text{MME}_{\text{C}}$. In POPE, we compute the average accuracy $\text{POPE}_{\text{Acc}}$ and F1-score $\text{POPE}_{\text{F1}}$ across three subsets: random, popular and adversarial.

\setlength{\tabcolsep}{5.2pt}
\begin{table*}[!t]
\begin{center}
\fontsize{7.pt}{7pt}\selectfont  % Set font size to 8.3pt and line spacing to 10pt
\begin{tabular}{lllrrrrrr}
\toprule
{Dataset}             & {Cost}                             & {Method}                         & \multicolumn{1}{c}{R Len}        & $\text{CHAIR}_{S}\downarrow$ & \multicolumn{1}{c}{$\text{CHAIR}_{I}\downarrow$} & \multicolumn{1}{c}{Recall$\uparrow$} & \multicolumn{1}{c}{Faith$\uparrow$} & \multicolumn{1}{c}{$\text{Faith}_{S}\uparrow$} \\ \midrule
\multicolumn{9}{c}{\cellcolor[HTML]{E5E5E5}\rule{0pt}{2.ex}LLaVA-v1.5-7b}   \\ [0.2ex]
                                         &                                                       & \rule{0pt}{2.3ex}{$\text{VCD}_{\text{t=500}}$}    & {101.3}         & 56.6                                             & 17.4                                             & 77.7                                 & 83.7                                & 67.1                                           \\
                                         &                                                       & {OPERA}                          & {96.5}          & \underline{52.6}                    & \underline{15.7}                    & 75.0                                 & \underline{85.9}       & \underline{71.0}                  \\
                                         & {\multirow{-3}{*}{Training-free}}  & {$\text{OPERA}_{\text{fast}}$}   & {102.8}         & 56.2                                             & 17.4                                             & 75.4                                 & 84.3                                & 68.6                                           \\ \cdashline{2-9}
{LLaVA-Details 23k}   &                                                       & \rule{0pt}{2.3ex}{MLE}                            & {102.1}         & 54.2                                             & 15.9                                             & 78.3                                 & \textbf{86.1}                       & \underline{71.0}                  \\
                                         &                                                       & {\textit{EOS}}                   & {\textit{62.6}} & \textit{30.2}                                    & \textit{10.9}                                    & \textit{69.2}                        & \textit{89.9}                       & \textit{74.7}                                  \\
                                         &                                                       & {\textbf{Ours $w_{\text{pos}}$}} & {102.5}         & 56.8                                             & 16.5                                             & \textbf{78.6}                        & 85.5                                & 70.2                                           \\
                                         & {\multirow{-4}{*}{Training-based}} & {\textbf{Ours $w_{\text{neg}}$}} & {101.8}         & \textbf{51.2}                                    & \textbf{15.5}                                    & \underline{78.4}        & 85.8                                & \textbf{72.3}                                  \\ \midrule
                                         &                                                       & {$\text{VCD}_{\text{t=200}}$}    & {105.2}         & 54.6                                             & 16.9                                             & 78.0                                 & 85.2                                & 71.2                                           \\
                                         &                                                       & {OPERA}                          & {97.2}          & \underline{51.2}                    & 14.9                                             & 76.4                                 & 86.1                                & 72.3                                           \\
                                         & {\multirow{-3}{*}{Training-free}}  & {$\text{OPERA}_{\text{fast}}$}   & {102.3}         & 54.6                                             & 15.6                                             & 77.3                                 & \underline{86.2}       & \textbf{72.6}                                  \\ \cdashline{2-9}
{LLaVA-Instruct 150k} &                                                       & \rule{0pt}{2.3ex}{MLE}                            & {105.3}         & 57.0                                             & 16.9                                             & \underline{78.7}        & 85.8                                & \underline{72.4}                  \\
                                         &                                                       & {\textit{EOS}}                   & {\textit{56.9}} & \textit{24.4}                                    & \textit{9.1}                                     & \textit{66.9}                        & \textit{90.9}                       & \textit{76.2}                                  \\
                                         &                                                       & {\textbf{Ours $w_{\text{pos}}$}} & {99.6}          & 52.2                                             & \underline{14.8}                    & \textbf{79.3}                        & 86.0                                & 71.8                                           \\
                                         & {\multirow{-4}{*}{Training-based}} & {\textbf{Ours $w_{\text{neg}}$}} & {99.8}         & \textbf{48.8}                                    & \textbf{13.7}                                    & 77.6                                 & \textbf{86.4}                       & 71.8                                           \\
\multicolumn{9}{c}{\cellcolor[HTML]{E5E5E5}\rule{0pt}{2.ex}LLaVA-v1.5-13b}   \\ [0.2ex]
                                         &                                                       & \rule{0pt}{2.3ex}{$\text{VCD}_{\text{t=200}}$}    & {104.4}         & 54.8                                             & 16.2                                             & 75.7                                 & 84.9                                & 69.5                                           \\
                                         &                                                       & {OPERA}                          & {97.5}          & 54.0                                             & 16.8                                             & 76.4                                 & 85.3                                & 70.0                                           \\
                                         & {\multirow{-3}{*}{Training-free}}  & {$\text{OPERA}_{\text{fast}}$}   & {98.8}          & 54.8                                             & 16.9                                             & 74.8                                 & 84.8                                & 68.8                                           \\ \cdashline{2-9}
{LLaVA-Details 23k}   &                                                       & \rule{0pt}{2.3ex}{MLE}                            & {105.2}         & 54.6                                             & \underline{15.3}                    & \underline{77.3}        & \textbf{86.6}                       & \underline{71.5}                  \\
                                         &                                                       & {\textit{EOS}}                   & {\textit{74.9}} & \textit{39.4}                                    & \textit{13.1}                                    & \textit{71.9}                        & \textit{88.8}                       & \textit{74.5}                                  \\
                                         &                                                       & {\textbf{Ours $w_{\text{pos}}$}} & {105.1}         & \underline{52.4}                    & 15.7                                             & \textbf{77.8}                        & 86.0                                & \textbf{71.8}                                  \\
                                         & {\multirow{-4}{*}{Training-based}} & {\textbf{Ours $w_{\text{neg}}$}} & {105.4}         & \textbf{51.0}                                    & \textbf{15.2}                                    & 76.6                                 & \underline{86.4}       & \textbf{71.8}                                  \\
\multicolumn{9}{c}{\cellcolor[HTML]{E5E5E5}\rule{0pt}{2.ex}PaliGemma-3b}    \\ [0.2ex]
                                         &                                                       & \rule{0pt}{2.3ex}{$\text{VCD}_{\text{t=500}}$}    & {98.9}          & 52.8                                             & 16.1                                             & 77.3                                 & 87.8                                & 71.9                                           \\
                                         &                                                       & {OPERA}                          & {102.3}         & 48.6                                             & \underline{13.5}                    & \underline{77.9}        & 87.5                                & 73.0                                           \\
                                         & {\multirow{-3}{*}{Training-free}}  & {$\text{OPERA}_{\text{fast}}$}   & {101.5}         & \underline{48.4}                    & \textbf{13.3}                                    & 76.7                                 & 88.0                                & \textbf{74.4}                                  \\ \cdashline{2-9}
{LLaVA-Details 23k}   &                                                       & \rule{0pt}{2.3ex}{MLE}                            & {100.7}         & 51.0                                             & 15.3                                             & 76.3                                 & \textbf{88.4}                       & \underline{74.0}                  \\
                                         &                                                       & {\textit{EOS}}                   & {\textit{42.2}} & \textit{15.8}                                    & \textit{6.5}                                     & \textit{56.1}                        & \textit{93.6}                       & \textit{80.4}                                  \\
                                         &                                                       & {\textbf{Ours $w_{\text{pos}}$}} & {100.5}         & 52.8                                             & 15.6                                             & \textbf{79.1}                        & 87.9                                & 73.1                                           \\
                                         & {\multirow{-4}{*}{Training-based}} & {\textbf{Ours $w_{\text{neg}}$}} & {101.0}         & \textbf{47.8}                                    & 14.6                                             & 74.4                                 & \underline{88.2}       & 73.7                                           \\
\multicolumn{9}{c}{\cellcolor[HTML]{E5E5E5}\rule{0pt}{2.ex}Bunny-v1.1-8b}   \\ [0.2ex]
                                         &                                                       & \rule{0pt}{2.3ex}{$\text{VCD}_{\text{t=200}}$}    & {343.6}         & 47.8                                             & 9.1                                              & 74.2                                 & 87.3                                & 72.7                                           \\
                                         &                                                       & {OPERA}                          & {334.1}         & \textbf{40.2}                                    & \textbf{7.7}                                     & 74.0                                 & \textbf{89.2}                       & 75.1                                           \\
                                         & {\multirow{-3}{*}{Training-free}}  & {$\text{OPERA}_{\text{fast}}$}   & {347.6}         & 42.4                                             & 8.8                                              & 73.9                                 & \underline{88.8}       & 74.2                                           \\ \cdashline{2-9}
{SVIT-Details 71k}    &                                                       & \rule{0pt}{2.3ex}{MLE}                            & {350.1}         & 44.4                                             & 8.7                                              & 74.4                                 & 87.8                                & 75.2                                           \\
                                         &                                                       & {EOS}                            & {311.9}         & 40.8                                             & \underline{8.2}                    & \underline{74.5}        & 88.2                                & \textbf{76.0}                                  \\
                                         &                                                       & {\textbf{Ours $w_{\text{pos}}$}} & {343.8}         & 44.0                                             & 8.5                                              & \textbf{75.9}                        & 88.6                                & \underline{75.4}                  \\
                                         & {\multirow{-4}{*}{Training-based}} & {\textbf{Ours $w_{\text{neg}}$}} & {353.4}         & \underline{40.4}                    & 8.5                                              & 73.5                                 & \underline{88.8}       & 75.1                                           \\ \bottomrule
\end{tabular}
\end{center}
\caption{Main results of three LVLMs on CHAIR and FaithScore. R Len represents the average response length. \textit{The rows in italics} are excluded from the comparison because their average response length is significantly lower than the others, making the comparison unfair. The best results are marked as \textbf{bold} and the second best are \underline{underlined}.}
\label{table:main_results}
\end{table*}

\begin{table}[t]
\begin{center}
\fontsize{7pt}{6pt}\selectfont
\begin{tabular}{llcccc}
\toprule
Dataset                                & Method                & $\text{MME}_{\text{P}}\uparrow$ & $\text{MME}_{\text{C}}\uparrow$ & $\text{POPE}_{\text{Acc}}\uparrow$ & $\text{POPE}_{\text{F1}}\uparrow$ \\ \midrule
\multirow{3}{*}{\begin{tabular}[c]{@{}l@{}}LLaVA-Instruct 150k\end{tabular}}   & MLE                   & 756.9  & 261.8  & 59.9      & 71.2     \\
                                       & EOS                   & 840.8  & 269.6  & 62.8      & 72.6     \\
                                       & Ours $w_{\text{neg}}$ & \textbf{849.9}  & \textbf{271.4}  & \textbf{64.2}      & \textbf{73.4}     \\ \midrule
\multirow{3}{*}{\begin{tabular}[c]{@{}l@{}}LLaVA-Mix Subset 264k\end{tabular}} & MLE                   & 1377.4 & 268.2  & 85.4      & 84.9     \\
                                       & EOS                   & 1369.1 & 267.9  & \textbf{85.9}      & \textbf{85.3}     \\
                                       & Ours $w_{\text{neg}}$ & \textbf{1386.6} & \textbf{282.1}  & \textbf{85.9}      & \textbf{85.3}     \\
                                       \bottomrule
\end{tabular}
\end{center}
\caption{The performance comparison on MME and POPE.}
\label{table:mme_pope}
\end{table}

\subsection{Baselines}

We select \textbf{two training-free} baselines and \textbf{two training-based} baselines. \textbf{OPERA} \citep{opera}: This inference-stage method identifies summary tokens that contribute to hallucinations and penalizes them. It further removes these tokens and re-samples the remaining ones to improve  response accuracy. \textbf{VCD} \citep{vcd}: This decoding strategy mitigates the model's language bias by contrasting the output distribution with an additional distribution generated from noisy images. \textbf{MLE}: This is the standard supervised training method using maximum likelihood estimation without any modifications. \textbf{EOS} \citep{eos}: This training-phase method excludes EOS (End of Sequence) token probabilities from the loss calculation for non-EOS tokens, allowing the model to terminate generation before producing redundant, often hallucinatory, content.  We perform greedy sampling except for OPERA which requires beam search.

\subsection{Main Results}

We conduct experiments by fine-tuning the pre-trained LVLMs with LoRA \citep{hu2021lora} for 2 epochs with batch size of 128. The inference-based methods are applied to models fine-tuned using a standard maximum likelihood estimation loss function. For our method, we fix the temperature $\tau$ at $0.5$ and apply the loss re-weighting only after 1 epoch to allow the model to warm-up and assess visual dependence more accurately. The noise time step is fixed to 900. \textbf{Table \ref{table:main_results}} presents the response length and the performance on CHAIR and FaithScore.

\paragraph{Emphasizing image-positive tokens increases both object recall and hallucination.} As discussed in \textbf{Section \ref{sec:hallucination_in_text_tokens}}, image-positive tokens contribute to both object extractions and hallucinations. Therefore, up-scaling the loss for image-positive tokens enables generating richer responses, but also increases hallucinations. This strategy might be a good choice in scenarios where hallucination is more acceptable, and users require responses with richer visual information \citep{zhai2023halle}.

\paragraph{Emphasizing image-negative tokens reduces hallucination.} This improvement in mitigating hallucination can be attributed to two factors. First, image-negative tokens are typically non-object words and are often located farther from where hallucinations tend to occur, as shown in \textbf{Figure \ref{fig:inspect_chair_co_occur}}. Second, it simultaneously downscales the loss weights for image-invariant and image-positive tokens, which are more likely to contain hallucinations. Therefore, we effectively reduce the model's focus on noisy content. As a result, by directing the model's attention away from these tokens, we help prevent it from learning and generating hallucinated information.

\paragraph{VCD performs poorly and EOS generates short responses.} We test VCD algorithm using three noise time steps $T=200, 500, 900$ and report the best one. We find that it does not perform as well as expected. One possible reason is that VCD does not perform well on reducing hallucinations from image-invariant tokens as they are not affected by the image conditions. We also test EOS and find it is highly effective at reducing hallucinations, but it comes at the cost of shorter response lengths and lower object recall scores. Naturally, shorter responses contain fewer details and content, making them produce less hallucinations.

\subsection{Performance on MME and POPE}

We report performance comparison on MME and POPE in \textbf{Table \ref{table:mme_pope}}. We select LLaVa-Instruct 150k and a subset from LLaVA-Mix 665k dataset \citep{llava} which consists of AOKVQA (66k), RefCOCO (48k), LLaVA-Conversation (58k), VQAv2 (83k) and OKVQA (9k), to save computational cost. This subset has 264k samples in total. We fine-tune LLaVA-v1.5-7b with both datasets on top of pre-trained checkpoints and evaluate their performance. Overall, our method is better on both MME and POPE when fine-tuned with LLaVA-Instruct 150k. Moreover, our method shows better results than EOS on MME and same performance on POPE as EOS trained with LLaVa-Mix Subset. One possible explanation is that the responses in LLaVA-Mix subset 264k are usually few tokens long, which are much shorter than those in LLaVa-Instruct 150k, leaving less space for loss adjusting.

\begin{table}[!t]
\begin{center}
\fontsize{7pt}{6pt}\selectfont
\begin{tabular}{llrccc}
\toprule
Model                           & Method                       & Length & $\text{CHAIR}_{S}\downarrow$  & $\text{CHAIR}_{I}\downarrow$  & Recall$\uparrow$              \\ \midrule
\multirow{7}{*}{LLaVA-v1.5-7b}  & -                            & 101.4  & 51.4                          & 14.0                          & \textbf{78.7}                 \\
                                & MLE                          & 100.4  & 48.8                          & 13.8                          & 77.9                          \\
                                & EOS                          & 82.9   & \textbf{41.0}                 & \textbf{12.1}                 & 75.0                          \\
                                & $\text{VCD}_{\text{t=500}}$  & 101.5  & 52.8                          & 14.8                          & \underline{78.6} \\
                                & OPERA                        & 93.1   & 42.8                          & 13.1                          & 76.9                          \\
                                & $\text{OPERA}_{\text{fast}}$ & 97.2   & 48.2                          & 13.7                          & 75.7                          \\
                                & \textbf{Ours $w_{\text{neg}}$}        & 97.9   & \underline{42.4} & \underline{12.6} & 76.5                          \\ \midrule
\multirow{7}{*}{LLaVA-v1.5-13b} & -                            & 102.2  & 48.4                          & 13.4                          & 77.4                          \\
                                & MLE                          & 101.0  & 48.2                          & 14.2                          & \textbf{77.8}                 \\
                                & EOS                          & 85.7   & \textbf{41.8}                 & 12.9                          & 75.1                          \\
                                & $\text{VCD}_{\text{t=200}}$  & 102.0  & 51.8                          & 14.3                          & \underline{77.5} \\
                                & OPERA                        & 102.4  & 43.3                          & \underline{12.7} & 76.8                          \\
                                & $\text{OPERA}_{\text{fast}}$ & 98.7   & 47.5                          & 13.6                          & 76.9                          \\
                                & \textbf{Ours $w_{\text{neg}}$}        & 99.1   & \underline{42.4} & \textbf{12.2}                 & 76.5                          \\ \bottomrule
\end{tabular}
\end{center}
\caption{The experimental results of continued fine-tuning of instructional tuned models.}
\label{table:continue_finetuning}
\end{table}

\subsection{Further Training Fine-tuned LVLMs}

Another experiment is conducted where we further train the already fine-tuned LLaVa-v1.5 7b and 13b on LLaVa-Detail 23k with LoRA for an additional epoch with batch size of 128. Given the shorter training duration, a larger temperature of $\tau=1$ is used. Re-weighting of the loss begins after half an epoch of training. The performance results are shown in \textbf{Table \ref{table:continue_finetuning}}. In the table, the method labeled with '-' represents the original instructional fine-tuned baseline model. Our approach, which suppresses image-invariant and image-positive tokens during training, achieves a notable reduction in hallucinations while preserving response length. Although EOS achieves the lowest hallucination rate, it does so at the expense of producing shorter responses. Moreover, as with most existing hallucination reduction methods, our approach leads to a decrease in recall, since the model becomes less inclined to identify additional objects in order to minimize the risk of hallucinations.

\subsection{Filtering Data by Visual Dependence}

We design a data filtering strategy by ranking each sample according to their total visual dependence, calculated as $\sum_{i=1}^T d({y_t})$. The experiment is conducted using LLaVA-Instruct 150k dataset, with visual dependence evaluated by the fine-tuned LLaVA-v1.5-7b model and the noise step is set at 900. We fine-tune pretrained LLaVA-v1.5-7b from scratch using filtered dataset for 2 epochs with LoRA, employing the vanilla loss function. \textbf{Table \ref{table:data_filtering}} presents the results. Notably, our filtering strategy does not affect the overall training data length or the response length. On the one hand, removing the top 10\% of data with the highest visual dependence significantly reduces hallucinations. However, this strategy lowers recall score, as image-positive data, which contain both valuable visual information and noisy hallucinated information, are removed. On the other hand, removing data with the lowest visual dependence helps control hallucinations without compromising recall scores, since these data contain fewer visual details compared to those with high visual dependence.

\begin{table}[t]
\begin{center}
\fontsize{7pt}{6pt}\selectfont
\begin{tabular}{lcrccc}
\toprule
Strategy & D Len & R Len & $\text{CHAIR}_{S}\downarrow$ & $\text{CHAIR}_{I}\downarrow$ & Recall$\uparrow$\\ \midrule
No Filtering          & 172.8    & 105.3        & 57.0     & 16.9     & \textbf{78.7}   \\ \midrule
Random 10\%        & 172.7    & 101.4        & 52.8     & 15.3     & \underline{78.6}   \\
Random 20\%        & 172.8    & 99.3         & 50.4     & 14.7     & 78.3   \\ \midrule
Lowest 10\%      & 172.9    & 99.6         & 49.2     & 14.2     & 78.3   \\
Lowest 20\%      & 172.4    & 100.2        & \underline{46.8}     & \underline{13.3}     & 78.4   \\ \midrule
Highest 10\%     & 176.1    & 100.8        & \textbf{44.4}     & \textbf{12.9}     & 76.9   \\
Highest 20\%     & 177.8    & 102.5        & 48.4     & 14.5     & 75.2   \\ \bottomrule
\end{tabular}
\end{center}
\caption{The experimental results of data filtering. D Len represents the average data sentence length of filtered data, and R Len represents the average response length.}
\label{table:data_filtering}
\end{table}

\section{Conclusion}

This work investigates the relationship of object hallucination with visual dependence of generated tokens. Our findings suggest that hallucinations occur around image-invariant tokens are mostly affected by the language context during generation. Moreover, image-positive tokens contribute to both hallucinations and extraction of visual information. Based on these findings, we develop a loss re-weighting approach and a data filtering strategy to mitigate hallucinations in LVLMs.

\bibliography{egbib}
\bibliographystyle{iclr2025_conference}

%=========================================================
% APPENDIX
%=========================================================
\appendix

\section{Additional Studies}
\label{sec:additional_studies}

\subsection{Impact of Different Hyperparameters}

We investigate how various hyperparameters influence  hallucination performance, including temperature $\tau$ in Equation~3, the starting time for loss adjustment, and the noise time steps used for calculating visual dependence. The results for each hyperparameter are detailed in \textbf{Table \ref{table:tau}, \ref{table:start_time}, \ref{table:noise_step}}, respectively. These experiments incorporate loss re-weighting, emphasizing image-negative tokens to reduce hallucinations. The default settings are $\tau$ of 0.5, adjustment starting after 50\% of the total training time, and a noise step $T$ of 900.

\paragraph{Temperature $\tau$} When the temperature $\tau$ is set to 0, it becomes the vanilla loss without any re-weighting. As $\tau$ becomes larger, the distribution of loss weights becomes sharper, meaning that the weights are more concentrated to emphasize only few image-negative tokens. As shown in \textbf{Table \ref{table:tau}}, the recall score drops with higher temperatures because the image-invariant and image-positive tokens are more heavily suppressed.

\paragraph{Start Time for Weight Adjustment} Since our loss adjustment approach requires assessing the visual dependence of each token, applying this adjustment too early can lead to inaccurate assessments and negatively impact performance. As shown in \textbf{Table \ref{table:start_time}}, starting the loss adjustment after the model has completed  half (50\%) of its total training time yields  better results than starting at the beginning (0\%).

\paragraph{Noise Step $T$} Different noise steps $T$ introduce varying levels of visual uncertainty when computing visual dependence, thereby affecting the calculation. A lower noise step generally produces visual dependence values closer to 0, whereas a higher noise step results in values with a larger magnitude. As \textbf{Table \ref{table:noise_step}} shows, a $T$ value at 200 provides the best recall performance, and a $T$ value at 900 provides the best hallucination reduction performance. Moreover, the LLaVA model is less sensitive to $T$ compared to the PaliGemma model.

Based on these results, for a better trade-off between hallucination and recall score, we choose a temperature $\tau$ of 0.5, adjustment starting after 50\% of the total training time, and a noise step $T$ of 900 as our default settings.

\begin{table}[!t]
\begin{center}
\fontsize{8pt}{8pt}\selectfont
\begin{tabular}{ccccc}
\toprule
{Temp $\tau$} & R Len & $\text{CHAIR}_{S}\downarrow$ & $\text{CHAIR}_{I}\downarrow$ & Recall$\uparrow$ \\ \midrule
\multicolumn{5}{c}{LLaVA-v1.5-7b  + LLaVA-Details 23k}                        \\ \midrule
{0.25}               & 101.9 & 52.2     & 15.9     & 77.9   \\
{0.50}               & 101.8 & \textbf{51.2}     & \textbf{15.5}     & \textbf{78.4}   \\
{1.00}               & 100.7 & 52.0     & 15.8     & 77.0   \\
{1.50}               & 101.1 & 51.8     & 16.1     & 75.9   \\ \midrule
\multicolumn{5}{c}{PaliGemma-3b  + LLaVA-Details 23k}                          \\ \midrule
{0.25}               & 100.8 & 49.8     & 15.1     & \textbf{76.4}   \\
{0.50}               & 101.0 & 47.8     & 14.6     & 74.4   \\
{1.00}               & 100.4 & 46.8     & 14.1     & 74.5   \\
{1.50}               & 100.4 & \textbf{44.4}     & \textbf{13.8}     & 74.0   \\ \bottomrule
\end{tabular}
\end{center}
\caption{The performance on CHAIR with different temperature $\tau$ in fine-tuning pre-trained LLaVA-v1.5-7b and PaliGemma-3b with LLaVA-Details 23k dataset.}
\label{table:tau}
\end{table}

\begin{table}[!t]
\begin{center}
\fontsize{8pt}{8pt}\selectfont
\begin{tabular}{ccccc}
\toprule
{Start Time} & R Len    & $\text{CHAIR}_{S}\downarrow$ & $\text{CHAIR}_{I}\downarrow$ & Recall$\uparrow$ \\ \midrule
\multicolumn{5}{c}{LLaVA-v1.5-7b +   LLaVA-Details 23k}                \\ \midrule
{0\%}        & 102.1 & 51.4     & 15.4     & 77.7   \\
{25\%}       & 102.3 & 52.2     & \textbf{15.3}     & 77.6   \\
{50\%}       & 101.8 & \textbf{51.2}     & 15.5     & \textbf{78.4}   \\
{75\%}       & 101.5 & 53.4     & 16.1     & 77.6   \\ \midrule
\multicolumn{5}{c}{PaliGemma-3b  + LLaVA-Details 23k}                  \\ \midrule
{0\%}        & 100.7 & 48.4     & 14.9     & \textbf{76.0}   \\
{25\%}       & 101.3 & 50.8     & 15.0     & 75.6   \\
{50\%}       & 101.0 & 47.8     & \textbf{14.6}     & 74.4   \\
{75\%}       & 101.0 & \textbf{47.6}     & 14.8     & 75.4   \\ \bottomrule
\end{tabular}
\end{center}
\caption{The performance on CHAIR with different start times of loss re-weighting in fine-tuning pre-trained LLaVA-v1.5-7b and PaliGemma-3b with LLaVA-Details 23k dataset. For example, 0\% means that starting loss re-weighting from the beginning.}
\label{table:start_time}
\end{table}

\begin{table}[!t]
\begin{center}
\fontsize{8pt}{8pt}\selectfont
\begin{tabular}{ccccc}
\toprule
{Noise Step $T$} & R Len & $\text{CHAIR}_{S}\downarrow$ & $\text{CHAIR}_{I}\downarrow$ & Recall$\uparrow$ \\ \midrule
\multicolumn{5}{c}{LLaVA-v1.5-7b + LLaVA-Details 23k}                    \\ \midrule
{200}          & 101.9 & 51.2     & \textbf{15.1}     & \textbf{78.9}   \\
{500}          & 102.0 & \textbf{50.6}     & 15.2     & 78.4   \\
{900}          & 101.8 & 51.2     & 15.5     & 78.4   \\ \midrule
\multicolumn{5}{c}{PaliGemma-3b + LLaVA-Details 23k}                  \\ \midrule
{200}          & 100.7 & 50.8     & 15.1     & \textbf{77.0}   \\
{500}          & 100.8 & 50.6     & 15.3     & 76.1   \\
{900}          & 101.0 & \textbf{47.8}     & \textbf{14.6}     & 74.4   \\ \bottomrule
\end{tabular}
\end{center}
\caption{The performance on CHAIR with different noise step $T$ in fine-tuning pre-trained LLaVA-v1.5-7b and PaliGemma-3b with LLaVA-Details 23k dataset.}
\label{table:noise_step}
\end{table}

\begin{table*}[!htb]
\begin{center}
\fontsize{8pt}{8pt}\selectfont
\begin{tabular}{lcccc}
\toprule
Hyper-parameters            & LLaVA-v1.5-7b & {LLaVA-v1.5-13b} & {PaliGemma-3b} & {Bunny-v1.1-8b} \\ \midrule
Random seed                 & 42                                & 42                                 & 42                               & 42                                \\ \midrule
LoRA rank                   & 128                               & 128                                & 64                               & 128                               \\
LoRA alpha                  & 256                               & 256                                & 128                              & 256                               \\
LoRA dropout                & 0.05                              & 0.05                               & 0.05                             & 0.05                              \\ \midrule
Epochs                      & 2                                 & 2                                  & 2                                & 2                                 \\
Warmup ratio                & 0.03                              & 0.03                               & 0.03                             & 0.03                              \\ \midrule
Per device train batch size & 16                                & 16                                 & 16                               & 8                                 \\
Gradient accumulation step  & 2                                 & 2                                  & 2                                & 2                                 \\
GPU devices                 & 4                                 & 4                                  & 4                                & 4                                 \\
Global batch size           & 128                               & 128                                & 128                              & 64                                \\ \midrule
Deepspeed stage             & zero2                             & zero3                              & zero2                            & zero3                             \\
Precision                   & BFloat16                          & BFloat16                           & BFloat16                         & BFloat16                          \\ \midrule
Optimizer                   & AdamW                             & AdamW                              & AdamW                            & AdamW                             \\
Weight decay                & 0                                 & 0                                  & 0                                & 0                                \\
Base learning rate          & 2e-4                              & 2e-4                               & 1e-4                             & 2e-4                              \\
MM projector learning rate  & 2e-5                              & 2e-5                               & 1e-5                             & 2e-5                              \\ \bottomrule
\end{tabular}
\end{center}
\caption{The training settings for main experiments of fine-tuning LVLMs from scratch.}
\label{table:hyper_paramets}
\end{table*}

\setlength{\tabcolsep}{4.3pt}
\begin{table*}[!htb]
\begin{center}
\fontsize{8pt}{9pt}\selectfont
\begin{tabular}{l|ll|c|cc}
\toprule
Experiment         & Model          & Dataset             & Epochs & Vanilla (Hours) & Ours (Hours) \\ \midrule
From Scratch       & LLaVA-v1.5-7b  & LLaVA-Details-23k   & 2      & 0.9             & 1.1          \\
                   & LLaVA-v1.5-7b  & LLaVA-Instruct-150k & 2      & 9.7             & 11.2         \\
                   & LLaVA-v1.5-13b & LLaVA-Details-23k   & 2      & 1.7             & 2.0          \\
                   & PaliGemma-3b   & LLaVA-Details-23k   & 2      & 0.5             & 0.6          \\
                   & Bunny-v1.1-8b  & SVIT-Details-71k    & 2      & 6.8             & 8.0          \\ \midrule
Continue Fine-tune & LLaVA-v1.5-7b  & LLaVA-Details-23k   & 1      & 0.5             & 0.6          \\
                   & LLaVA-v1.5-13b & LLaVA-Details-23k   & 1      & 0.8             & 1.0          \\
                   & PaliGemma-3b   & LLaVA-Details-23k   & 1      & 0.2             & 0.3          \\
                   & Bunny-v1.1-8b  & SVIT-Details-71k    & 1      & 3.5             & 4.0          \\ \bottomrule
\end{tabular}
\end{center}
\caption{Computational cost for each experiment.}
\label{table:computational_cost}
\end{table*}

\section{Experiment Settings and Computational Cost}
\label{sec:exp_settings}

The experiments are conducted with 4 NVIDIA A100-40GB GPU cards. Due to heavy computational cost of training LVLMs, all experiments are single-run with random seed at 42. For reproducibility, we provide the detailed experiment settings for fine-tuning the pre-trained LVLMs in \textbf{Table \ref{table:hyper_paramets}}, including LoRA parameters, batch size, optimizer and etc. We use greedy decoding strategy for our method.

Our method introduces around 20\% extra training cost as we conduct evaluation of visual dependence during training. We present the computational cost of using the vanilla loss and our re-weighted loss in \textbf{Table \ref{table:computational_cost}}. Note that the visual dependence and loss weights is computed in evaluation mode, thus our method does not consume extra GPU memory.

\section{Data Filtering by Visual Dependence}

\begin{figure}[t]
    \begin{center}
        \includegraphics[width=0.678\linewidth]{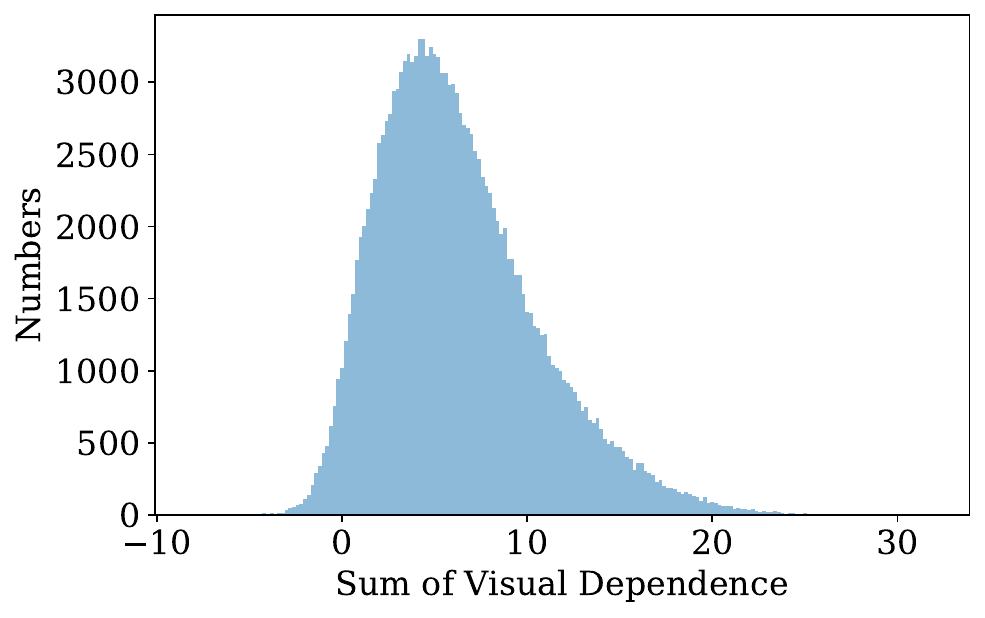}
    \end{center}
    \caption{The distribution of the sum of visual dependence of each sample from LLaVA-Instruct 150k.}
    \label{fig:visual_dependence}
\end{figure}

We remove data by ordering them of their visual dependence evaluated by an instructional fine-tuned model. \textbf{Figure \ref{fig:visual_dependence}} shows the metrics distribution of LLaVA-Instruct 150k dataset. The metrics is the sum of visual dependence of all tokens from each instruction $\sum_{t=1}^{T}d(y_t)$ evaluated by instructional fine-tuned LLaVA-v1.5-7b. A higher value indicates that the model's prediction probability will decrease, in other words, the model is less confident with its original response, after seeing a noisy image. As illustrated in the figure, most of instructions have positive values of visual dependence. By filtering out data with lowest value or with highest value, the hallucination performance of the model is affected in different ways.

\section{Additional Results on Further Training Fine-tuned LVLMs}

Here, we show more results of experiments on further training the already fine-tuned PaliGemma-3b with LLaVa-Detail 23k dataset and Bunny-v1.1-8b with SVIT-Detail 71k dataset with LoRA for an additional epoch with batch size of 128. The performance results are shown in \textbf{Table \ref{table:continue_finetuning_appendix}}. As shown in the table, our method achieves overall better performance on these two variants of LVLMs. Recall that our method also works on LLaVA series models. It demonstrates the general capability of our method.

\setlength{\tabcolsep}{4pt}
\begin{table}[!t]
\begin{center}
\fontsize{8pt}{9pt}\selectfont
\begin{tabular}{lcccc}
\toprule
{Method}       & {Length} & $\text{CHAIR}_{S}\downarrow$ & $\text{CHAIR}_{I}\downarrow$ & Recall$\uparrow$ \\ \midrule
\multicolumn{5}{c}{PaliGemma-3b +   LLaVA-Detail 23k}                                                                                  \\ \midrule
{\textit{-}}                   & {\textit{51.0}} & \textit{18.0} & \textit{7.0}  & \textit{58.1} \\
{MLE}                          & {66.1}          & 26.6          & 11.9          & \textbf{69.1} \\
{\textit{EOS}}                 & {\textit{33.4}} & \textit{13.2} & \textit{11.1} & \textit{57.8} \\
{$\text{VCD}_{\text{t=900}}$}  & {65.6}          & 29.6          & 10.6          & 68.6          \\
{OPERA}                        & {76.2}          & \underline{26.2}    & 8.8           & 67.8          \\
{$\text{OPERA}_{\text{fast}}$} & {82.7}          & \textbf{24.2} & \underline{8.7}     & \underline{69.0}    \\
{Ours $w_{\text{neg}}$}        & {71.3}          & \underline{26.2}    & \textbf{7.9}  & 68.6          \\ \midrule
\multicolumn{5}{c}{Bunny-v1.1-8b + SVIT-Detail 71k}                                                                                    \\ \midrule
{-}                            & {246.5}         & 40.6          & \underline{7.6}     & 74.3          \\
{MLE}                          & {300.8}         & 40.8          & \textbf{7.5}  & \textbf{75.6} \\
{EOS}                          & {272.5}         & 42.4          & 8.5           & 74.8          \\
{$\text{VCD}_{\text{t=200}}$}  & {300.5}         & 44.2          & 8.9           & 75.2          \\
{OPERA}                        & {293.4}         & \underline{37.8}    & 7.7           & 74.5          \\
{$\text{OPERA}_{\text{fast}}$} & {302.8}         & 39.4          & \textbf{7.5}  & \underline{75.4}    \\
{Ours $w_{\text{neg}}$}        & {302.7}         & \textbf{37.2} & 7.8           & 73.8          \\ \bottomrule
\end{tabular}
\end{center}
\caption{The additional experimental results of continued fine-tuning of instructional tuned models PaliGemma-3b and Bunny-v1.1-8b.}
\label{table:continue_finetuning_appendix}
\end{table}

\section{Qualitative Results}

\begin{figure*}[t]
    \begin{center}
    \includegraphics[width=0.98\linewidth]{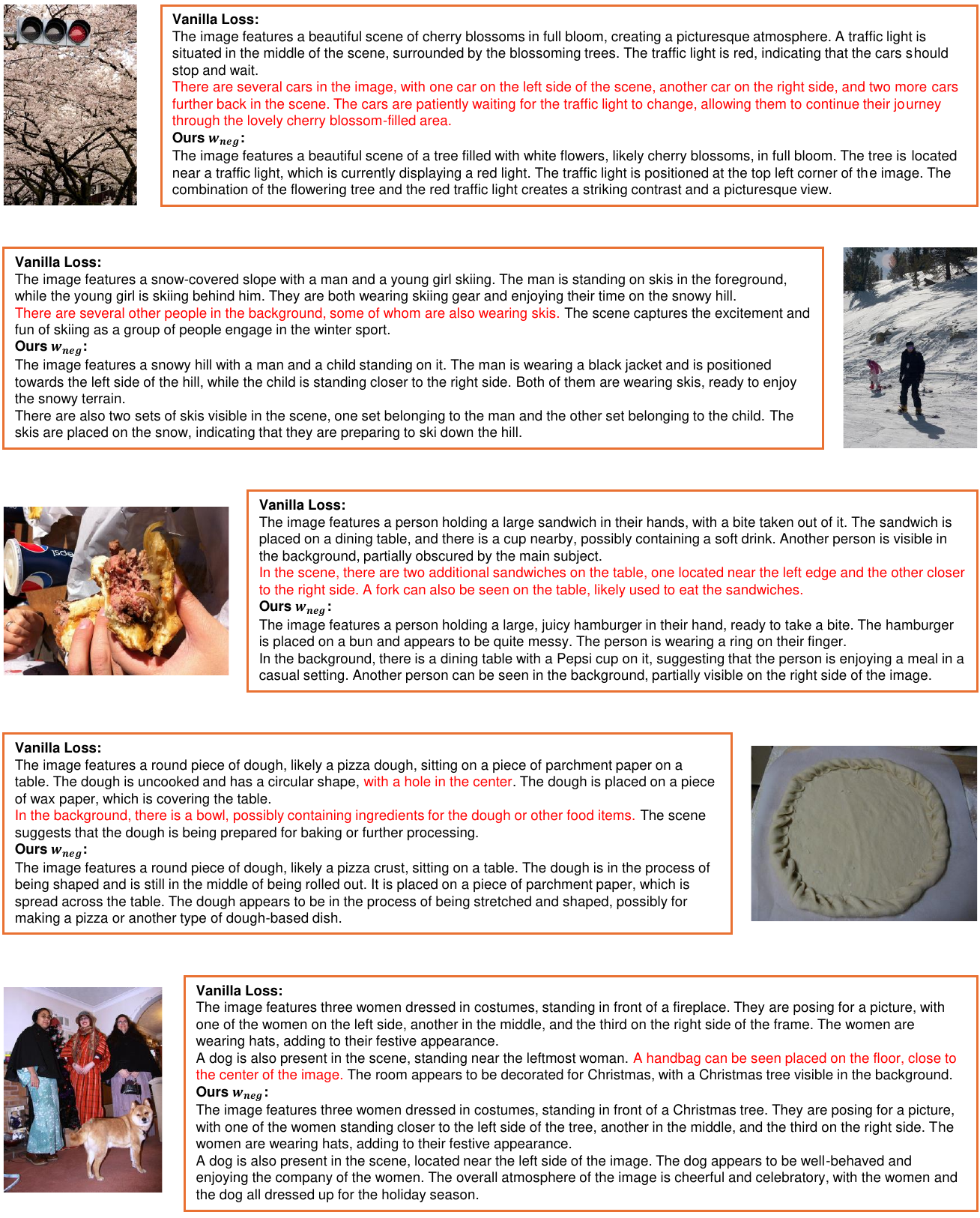}
    \end{center}
    \caption{Qualitative results of LLaVA-v1.5-7b trained with vanilla loss baseline and ours approach with  $w_{\text{neg}}$ on LLaVA-Instruct 150k dataset. Hallucinations are marked as red.}
    \label{fig:llava-7b-qualitative-results-page1}
\end{figure*}

\begin{figure*}[t]
    \begin{center}
    \includegraphics[width=0.98\linewidth]{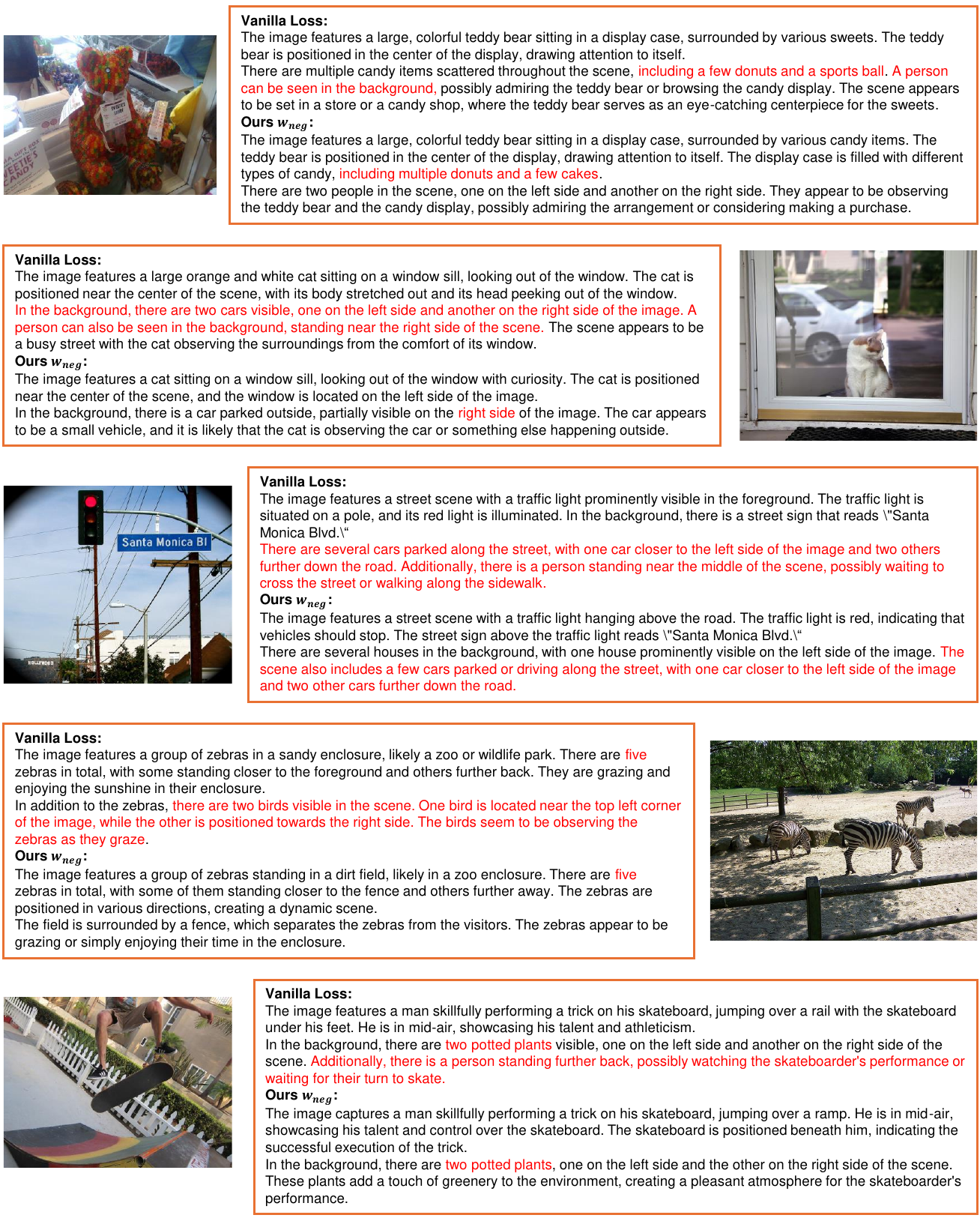}
    \end{center}
    \caption{Qualitative results of LLaVA-v1.5-7b trained with vanilla loss baseline and ours approach with  $w_{\text{neg}}$ on LLaVA-Instruct 150k dataset. Hallucinations are marked as red.}
    \label{fig:llava-7b-qualitative-results-page2}
\end{figure*}

\begin{figure*}[t]
    \begin{center}
    \includegraphics[width=0.98\linewidth]{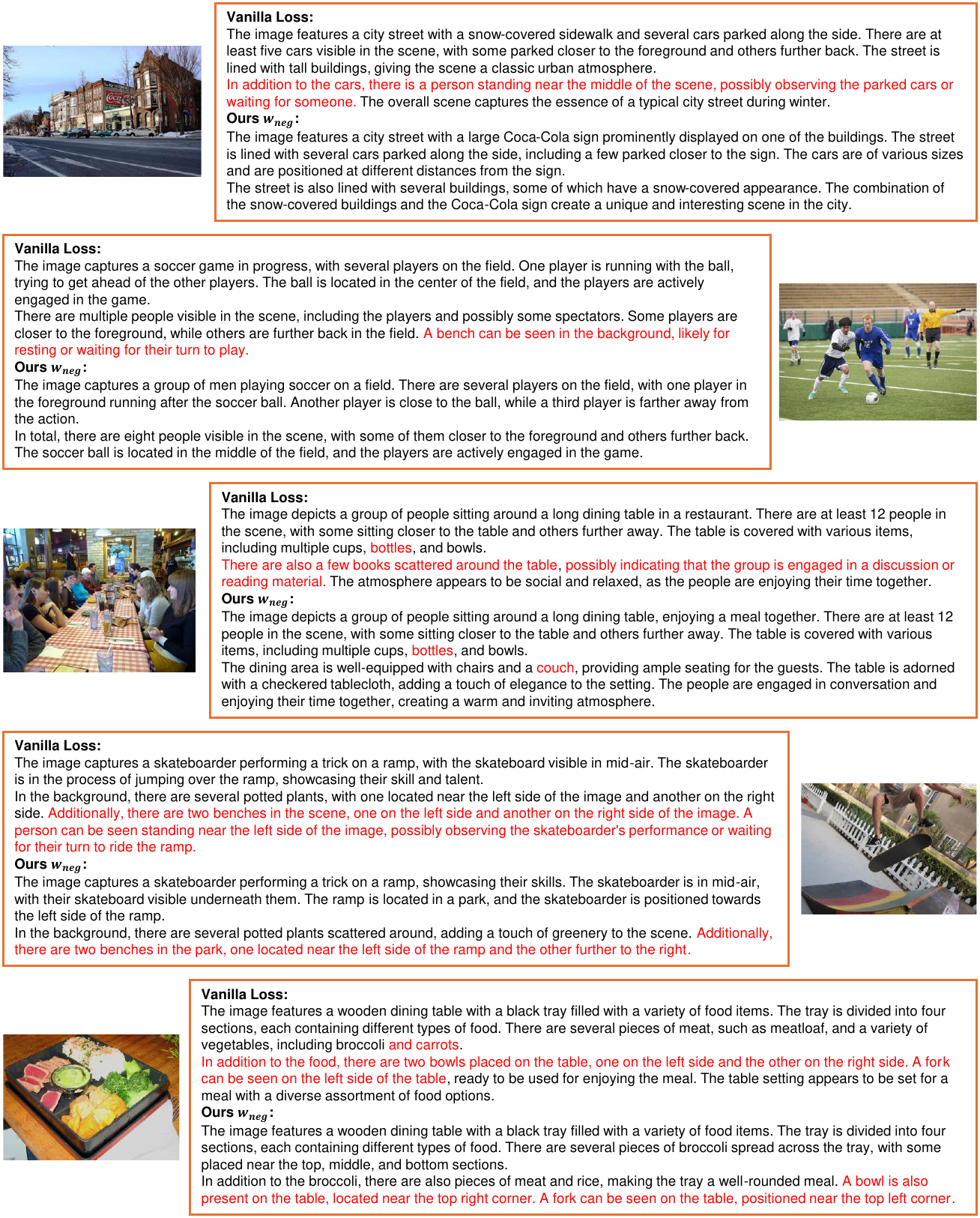}
    \end{center}
    \caption{Qualitative results of PaliGemma-3b trained with vanilla loss baseline and ours approach with  $w_{\text{neg}}$ on LLaVA-Details 23k dataset. Hallucinations are marked as red.}
    \label{fig:paligemma-qualitative-results-page1}
\end{figure*}

\begin{figure*}[t]
    \begin{center}
    \includegraphics[width=0.98\linewidth]{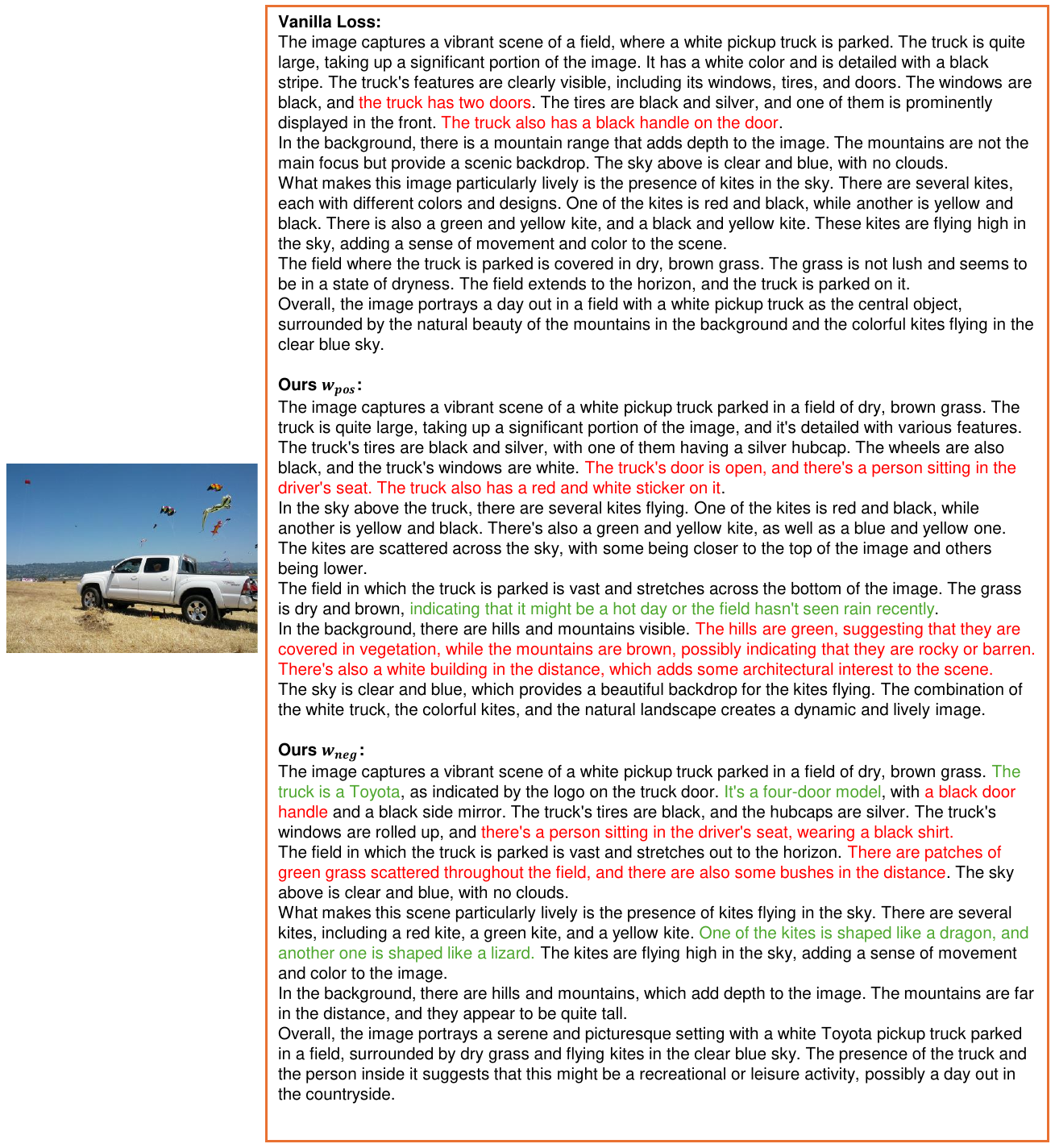}
    \end{center}
    \caption{Qualitative results of  Bunny-v1.1-8b trained with vanilla loss baseline and ours approach with $w_{\text{pos}}$ and $w_{\text{neg}}$ on SVIT-Details 71k dataset. Hallucinations are marked as red, the correct details that are not discovered in baseline are marked green.}
    \label{fig:bunny-qualitative-results-page1}
\end{figure*}

\begin{figure*}[t]
    \begin{center}
    \includegraphics[width=0.98\linewidth]{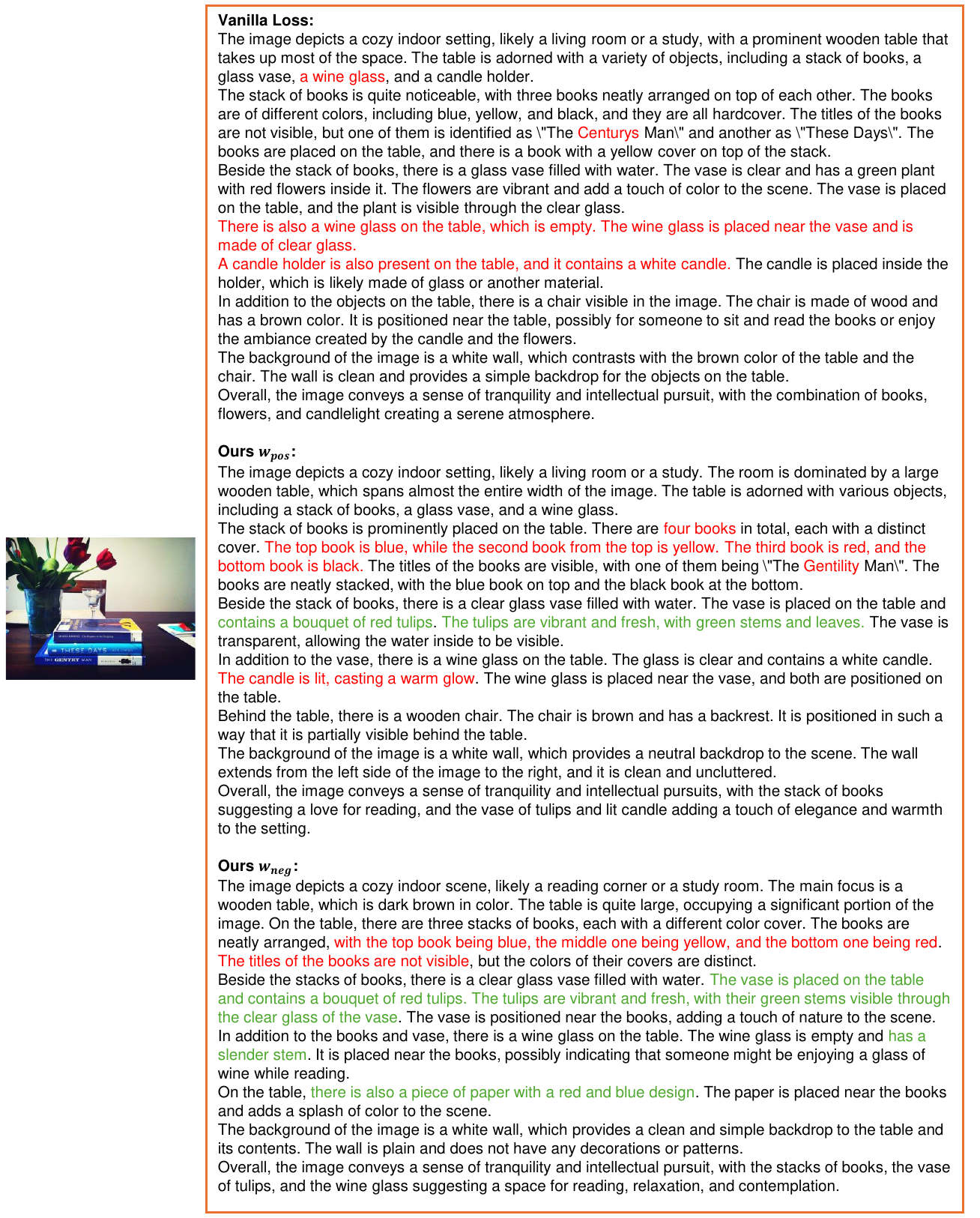}
    \end{center}
    \caption{Qualitative results of  Bunny-v1.1-8b trained with vanilla loss and ours approach with $w_{\text{pos}}$ and $w_{\text{neg}}$ on SVIT-Details 71k dataset. Hallucinations are marked as red, the correct details that are not discovered in baseline are marked green.}
    \label{fig:bunny-qualitative-results-page2}
\end{figure*}

\begin{figure*}[t]
    \begin{center}
    \begin{subfigure}[t]{\linewidth}
        \centering
        \includegraphics[width=1.0\linewidth]{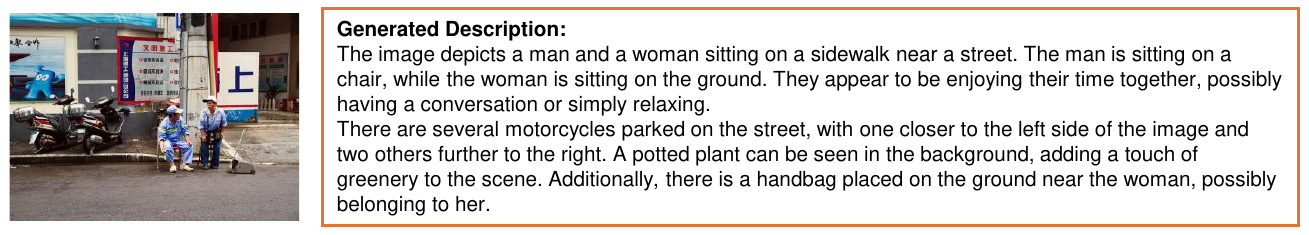}
    \end{subfigure}
    \begin{subfigure}[t]{\linewidth}
        \centering
        \includegraphics[width=1.0\linewidth]{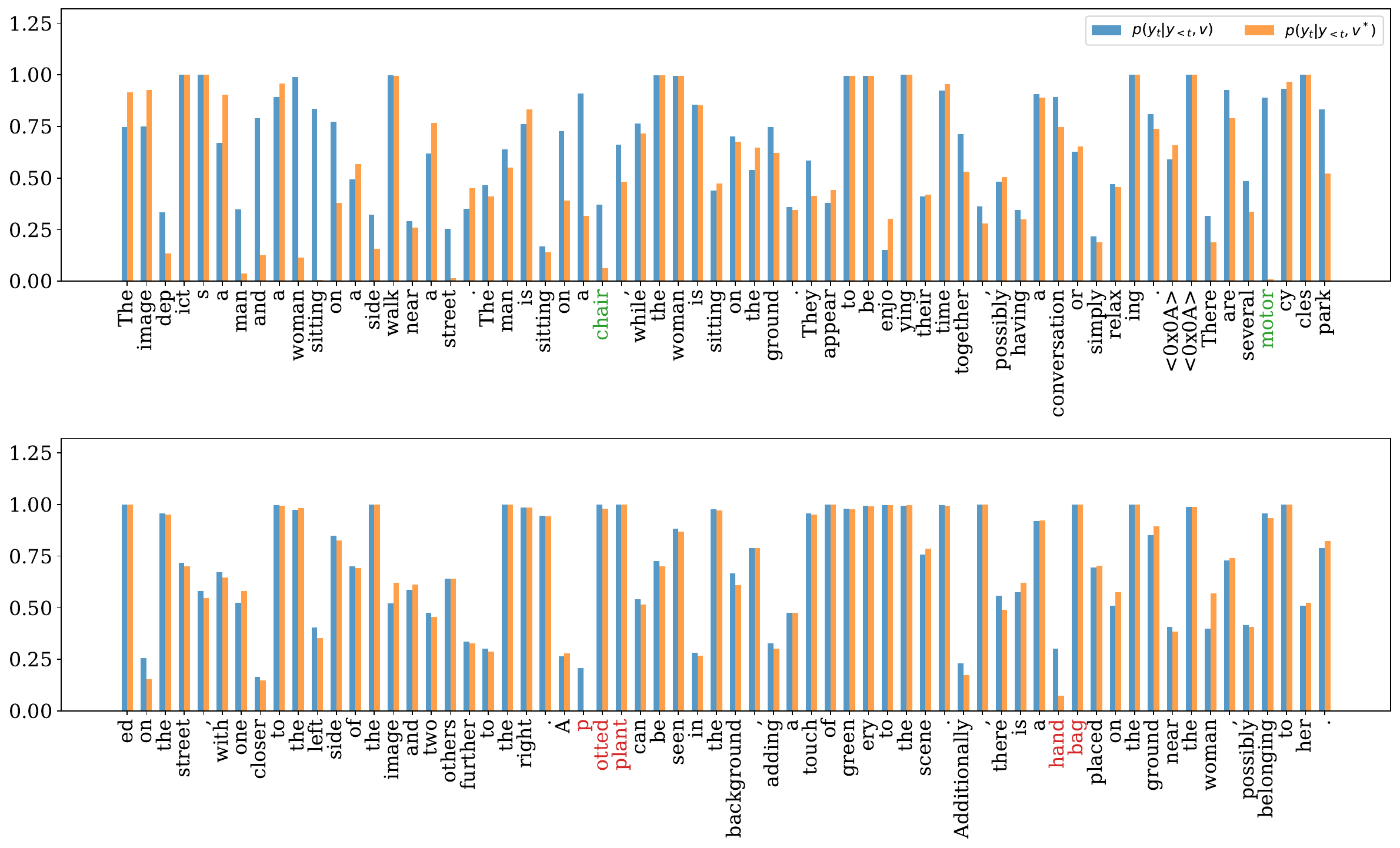}
    \end{subfigure}
    \end{center}
    \caption{The top left sub-figure is the input image, and the top right sub-figure is the generated image description from LLaVA-v1.5-7b fine-tuned from scratch with LLaVA-Details 23k dataset. The bottom bar charts display the probability of each token with clean image input $v$ and with noisy image input $v^*$. The hallucinated objects are marked red, the grounded objects are marked green.}
    \label{fig:prob_distribution_chair_8211}
\end{figure*}

\begin{figure*}[t]
    \begin{center}
    \begin{subfigure}[t]{\linewidth}
        \centering
        \includegraphics[width=1.0\linewidth]{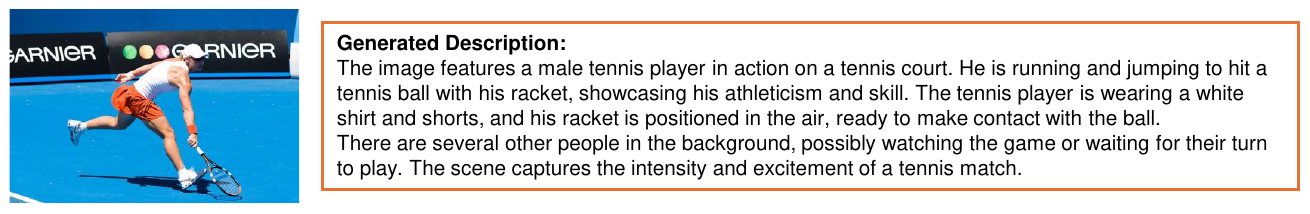}
    \end{subfigure}
    \begin{subfigure}[t]{\linewidth}
        \centering
        \includegraphics[width=1.0\linewidth]{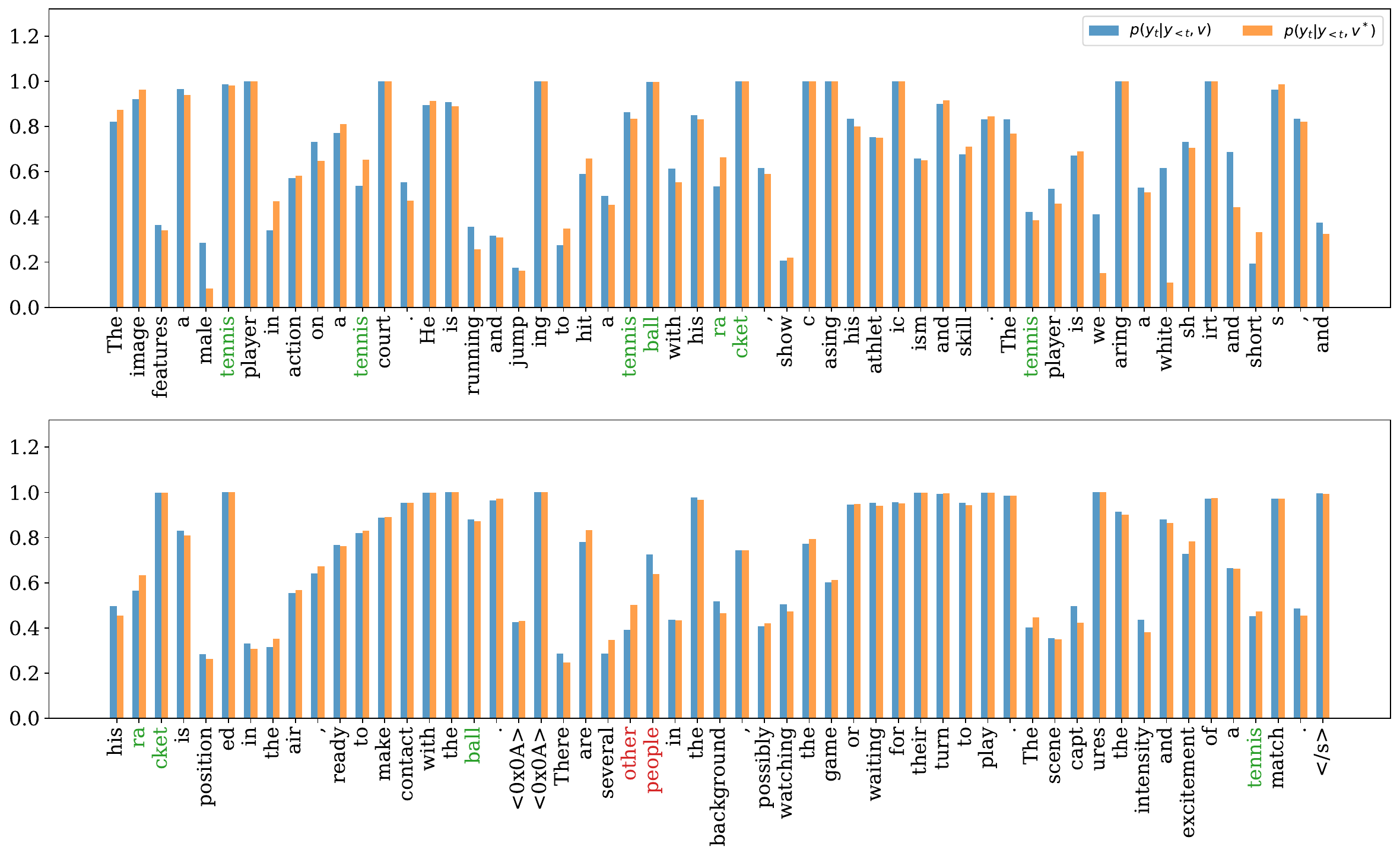}
    \end{subfigure}
    \end{center}
    \caption{The top left sub-figure is the input image, and the top right sub-figure is the generated image description from LLaVA-v1.5-7b fine-tuned from scratch with LLaVA-Details 23k dataset. The bottom bar charts display the probability of each token with clean image input $v$ and with noisy image input $v^*$. The hallucinated objects are marked red, the grounded objects are marked green.}
    \label{fig:prob_distribution_chair_112066}
\end{figure*}

We demonstrate qualitative results of generate image descriptions using the fine-tuned model from scratch with vanilla loss and our re-weighted loss by emphasizing image-negative tokens. \textbf{Figures \ref{fig:llava-7b-qualitative-results-page1}, \ref{fig:llava-7b-qualitative-results-page2}, \ref{fig:paligemma-qualitative-results-page1}, \ref{fig:bunny-qualitative-results-page1}, \ref{fig:bunny-qualitative-results-page2}} show the results. We show examples of probability distribution of predicted tokens with clean image and with noisy image of fine-tuned LLaVA-v1.5-7b, which is trained from scratch with LoRA on LLaVA-Details 23k dataset. The results are shown in \textbf{Figure \ref{fig:prob_distribution_chair_8211}, \ref{fig:prob_distribution_chair_112066}}.

\end{document}